\documentclass[12pt]{article}
\usepackage[a4paper, margin={1in}]{geometry}
\usepackage{graphicx}
\usepackage{cite}

\usepackage{multirow}
\usepackage{multicol}
\usepackage{booktabs}
\usepackage[pagebackref,breaklinks,colorlinks]{hyperref}

\newcommand\blfootnote[1]{%
  \begingroup
  \renewcommand\thefootnote{}\footnote{#1}%
  \addtocounter{footnote}{-1}%
  \endgroup
}

\begin{document}
\sloppy

\title{Fusing Pseudo Labels with Weak Supervision for Dynamic Traffic Scenarios}
\author{
    Anonymous ICCV BRAVO Workshop submission
}
\author{
    Harshith Mohan Kumar$^{1}$\thanks{Work done while an intern at Intel Corporation, hiharshith18@gmail.com},
    Sean Lawrence$^{2}$\thanks{Corresponding author, sean.j.lawrence@intel.com \\}
\\
$^{1}$Department of Computer Science, PES University\\
$^{2}$Intel Corporation}
\date{}

\maketitle

\blfootnote{This work was accepted as an extended abstract at the \emph{International Conference on Computer Vision (ICCV)} 2023 BRAVO Workshop, Paris, France.}

\begin{figure}[ht]
  \centering
  \includegraphics[width=0.8\textwidth]{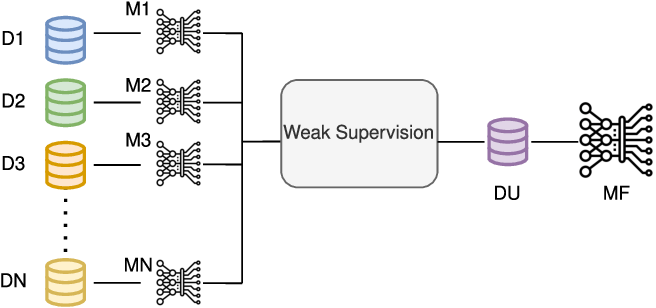}
  \caption{Proposed pipeline which merges pseudo labels from multiple models to form a unified dataset with weak supervision. The final model is trained on the unified label space.}
  \label{fig:pipeline}
\end{figure}

% ADAS Description + Role of computer vision in ADAS
Recent advancements in Advanced Driver Assistance Systems (ADAS) have witnessed remarkable progress and widespread adoption in the past decade. Historically, the automotive industry has been utilizing a fusion of various sensors, including lidar and radar~\cite{ADAS_Survey}. However, drastic improvements in computer vision networks have enabled improved perception, enhanced decision-making capabilities and accurate prediction of impending collisions from camera inputs alone. Moreover, the application of vision based deep learning models has enabled ADAS to learn from large datasets, improving their ability to recognize and interpret complex driving scenarios~\cite{ADAS_Trends}.

% CAS alert description
% Collision avoidance alerts are a critical component of ADAS which aid in preventing potential accidents by notifying drivers of potential hazards. These alert events include but are not limited to Forward Collision Warning (FCW), Pedestrian Collision Warning (PCW), Head-way Monitoring Warning (HMW) and Lane Departure Warning (LDW). The current state-of-the-art in Collision Avoidance Systems (CAS) leverage object detection, semantic segmentation and depth estimation to accurately predict collision probabilities.

% What is the problem?
While these systems have shown tremendous advancements, they face challenges in generalizing and adapting to various traffic conditions~\cite{ADAS_Perception}. One of the main limitations of camera based models arises from the fact that large popular datasets used for training these models indirectly introduces bias of location, weather and traffic patterns. Additionally, the driving conditions found in these datasets revolve around well-maintained road infrastructure~\cite{Comma2k,Kitti,NuScene,cityScape}. Changes in the environment, such as geographical locations, driving cultures and road infrastructure, can introduce distributional shifts~\cite{weather}. These limitations ultimately degrade performance and reduce potential safety risks in situations that differ from the training data.

% Summary of efforts to tackle this issue
Several semi-supervised techniques such as pseudo labeling~\cite{PseudoLabel}, contrastive learning~\cite{SimCLR,SimCLRv2}, and noisy student~\cite{NoisyStudent} have demonstrated ways to learn a model from limited annotations. However, these techniques have strong dependence on the initial labeled data, sensitive to label noise and have difficulty in handling concept drift.

% What is our solution?
In this work we propose a weakly-supervised label unification pipeline with pseudo labels to train a singular object detection model from multiple datasets. Our pipeline architecture is illustrated in Figure~\ref{fig:pipeline}. We initially fine-tune multiple homogeneous object detection models on each dataset. Then the final dataset $D_U$ spanning the entire label space is populated using pseudo labels. Labels which fall under a certain threshold are manually checked to avoid propagating errors from the initial model. Finally we retrain a singular object detection model on the combined label space to produce a robust model $M_F$ invariant to domain shifts. 

\paragraph{Label Unification Pipeline.} Following the work proposed in~\cite{multiDatasets}, we develop a label unification pipeline to combine $N$ heterogeneous datasets, $D_1,D_2,...D_N$ and corresponding their label spaces $L_1,L_2,...,L_N$. The label spaces may consist of non-disjoint sets, $L_i \cap L_j \neq \emptyset$. These labels are merged and verified through a human operator. This process allows us to train a single model with the union of all label spaces $L_U = L_1\cup L_2...\cup L_N$. 

To produce $L_U$ we initially fine-tune multiple detectors $M_1,M_2,...,M_N$ on each dataset. The architecture of the model used remains the same. Each model populates the other $N-1$ datasets with pseudo labels above a certain threshold. We do not adopt the custom loss function proposed in~\cite{multiDatasets}, instead labels which fall under the threshold are flagged and passed to a verification process where a human annotator validates the true label. We use the Intel Geti tool for visual inspection.

\paragraph{Dataset.}
To demonstrate that our model can work in adverse road environments, we choose to gather road facing images from countries across Asia. For object detection we chose to work with the Indian Driving Dataset (IDD)~\cite{IDD} and Road Damage Dataset (RDD)~\cite{RDD2022,RDD2020,RDD}. Additionally we manually procure a dataset containing over 2600 ten second video clips recorded at the occurrence of a collision avoidance alert triggered by a Mobileye 8 Connect device.

% \subsection{IDD}
% The Indian Driving Dataset \cite{IDD} consists of front facing images, of size 1920x1080, obtained from a camera attached to a car. The dataset is collected over the regions of the cities of Hyderabad and Bangalore.

% \subsection{RDD}
% The Road Damage Dataset contains of 47,420 road images from six different countries, Japan, India the Czech Republic, Norway, the United States, and China. The dataset has been annotated for various types of road damage. For our study we focus on extracting images consisting of crosswalk blur.

% \subsection{Weather-Time}
% For weather classification we obtain a dataset consisting of 2600 manually labeled weather and time labels from the China Computer Federation Big Data \& Computing Intelligence Contest (CCF BDCI). The images from this dataset are from the preliminary contest of CCF BDCI 2016 Traffic sign detection in self-driving scenarios. The original dataset comprises of 4,000 road images of size 1280Xx720. These images contain varieties in illumination and angles as they were picked up from a taxi driving recorder. The manually labeled weather categories include cloudy sunny and rainy.

\paragraph{Manual Alert Data.}
No public datasets consisting of Collision Avoidance Alert (CAS) alert metadata and scene frames are available for grading these alerts. To address this issue we develop a custom dataset that contains a diverse set of real-world driving scenes with various road types and collision scenarios. We generated over 2600 forward and pedestrian collision warnings across the city of Bangalore with varying lighting conditions.

% Alert Pipeline
% \begin{figure}
%     \centering
%        \includegraphics[width=0.8\linewidth]{images/AlertPipe.png}
%     \caption{Manual data collection pipeline.}
%     \label{fig:alertPipe}
% \end{figure} 

We use the Mobileye 8 Connect system which is a popular choice for numerous personal and commercial vehicles globally. This device is an advanced automotive vision-based platform designed to enhance road safety.

% At the time at which the alert was triggered by this device, we captured additional CAS alert metadata of alert type, date, time and location. Additionally, we obtain digital video recorder clips of ten seconds synchronized to the generation of the CAS alert by the Mobileye 8 Connect device.

\paragraph{Experiments.}
In this study we use two training procedures, one for training the object detection model and the other for weather classification. Our experiments were conducted on a system containing an Intel Core i5-9600K CPU paired with two Nvidia RTX 3060 GPUs with a total of 24GB of VRAM.

We train and compare performance across four popular object detection networks, namely YoloX~\cite{yolox}, Adaptive Sample Selection Training (ATSS)~\cite{ATSS}, Single Shot MultiBox Detector (SSD)~\cite{SSD}. We used a batch size of 64, learning rate of 0.01, weight decay of 0.005 and trained our models for 50 epochs. The models are trained using the combination of Binary Cross Entropy, Bounding Box Loss and Dual Focal Loss.

We compare the class wise F1 scores and the mean average precision (mAP) on the training set across all four networks in Table~\ref{tab:modelComparision}. Of the four networks, the Yolov8 model performs significantly better than the rest. This model was then trained again on a label unified dataset using pseudo labels with manual verification on less confident predictions. We present the final validation set class-wise results in Table \ref{tab:yolov8}.

\begin{table*}[t]
    \centering
    \begin{tabular}{c l c c c c c c c c c c}
        \toprule
        \multirow{2}{*}{\bfseries Model} & 
        \multirow{2}{*}{\bfseries Training} &
        \multicolumn{9}{c}{\bfseries F1 Per Label} &
        \multirow{2}{*}{\bfseries mAP (\%)}
        %------ Multicolumn headings
        \\ \cmidrule(lr){3-11}
            &&All&A&TS&M&R&P&C&Car&VF
        \\ \cmidrule(lr){1-12}
        %------
        YoloX & S & 0.42 & 0.49 & 0.19 & 0.45 & 0.38 & 0.24 & 0.69 & .55 & 0.04 & 0.32\\
        ATSS & S & 0.61 & 0.73 & 0.53 & 0.62 & 0.55 & 0.50 & 0.79 & 0.72 & 0.24 & 0.54\\
        SSD & S & 0.40 & 0.40 & 0.01 & 0.48 & 0.37 & 0.18 & 0.56 & 0.53 & 0.06 & 0.28\\
        Yolov8 & S & 0.60 & 0.75 & 0.73 & 0.69 & 0.68 & 0.54 & 0.80 & 0.79 & 0.32 & 0.65\\
        \hline
        \textbf{Yolov8*} & SSL & 0.77 & 0.85 & 0.82 & 0.79 & 0.75 & 0.69 & 0.89 & 0.83 & 0.42 & \textbf{0.78}\\
        \bottomrule
    \end{tabular}
    \caption{Training results on the four different object detection networks. * Indicates the model trained on pseudo labeled dataset. Please see text for more details on the training sets and the baselines.}
  \label{tab:modelComparision}
\end{table*}

\paragraph{Conclusion.} In this work we demonstrated the effectiveness of our weakly-supervised pseudo labeling pipeline in handling data distribution shifts. Our work can positively influence the accuracy of downstream ADAS tasks such as Collision Avoidance Alerts in areas with poor road infrastructure. In future work, we aim to demonstrate the capability of this pipeline for other computer vision tasks such as classification and segmentation.

\begin{table}[]
    \centering
    \begin{tabular}{c|c|c|c|c}
        \toprule
        Class  & Precision & Recall & mAP50 & mAP50-95\\
        \cmidrule(lr){1-5}
        all  & 0.887 & 0.677 & 0.808 & 0.613\\
        A  & 0.927 & 0.785 & 0.886 & 0.738\\
        TS & 0.887 & 0.766 & 0.866 & 0.657\\
        M & 0.892 & 0.711 & 0.839 & 0.614\\
        R & 0.907 & 0.638 & 0.796 & 0.574\\
        P & 0.840 & 0.585 & 0.744 & 0.521\\
        C & 0.896 & 0.874 & 0.91 & 0.689\\
        CA &  0.887 & 0.772 & 0.866 & 0.708\\
        VF &  0.857 & 0.281 & 0.554 & 0.404\\
        % TPVFormer \cite{huang2023triperspective} &  SemanticKITTI & 35.61\\
        % MonoScene \cite{cao2022monoscene} &  SemanticKITTI & 36.86\\
        % JS3C-Net \cite{yan2021sparse} &  SemanticKITTI  & \textbf{38.98}\\
        % AICNet \cite{li2020anisotropic} &  SemanticKITTI & 29.59\\
        % 3DSketch \cite{chen20203d} & SemanticKITTI  & 33.30\\
        % LMSCNet \cite{roldão2020lmscnet} &  SemanticKITTI & 28.61\\
        % DEST-B3 \cite{TODO} & CityScapes & \textbf{72.58}\\
        % SegFormer-B3 \cite{TODO} & CityScapes & 72.30\\
        \bottomrule
    \end{tabular}
    \caption{Validation results of the chosen Yolov8 model ($M_F$) after training on pseudo labels.}
    \label{tab:yolov8}
\end{table}

\bibliographystyle{IEEEtran}
\bibliography{references}

\end{document}